\documentclass[11pt]{article}

\usepackage[final]{acl}

\usepackage{times}
\usepackage{latexsym}

\usepackage[T1]{fontenc}
\usepackage[utf8]{inputenc}

\usepackage{microtype}

\usepackage{inconsolata}

\usepackage{booktabs}
\usepackage{subcaption}
\usepackage{multirow}
\usepackage{tabularx}
\usepackage{natbib}
\usepackage{graphicx}
\usepackage{amssymb}
\usepackage{threeparttable}

\title{Isolated Sign Language Recognition for Icelandic Sign Language: Experiments in a Low-resource Setting}

\author{Finnur Ágúst Ingimundarson$^1$, Guðný Björk Þorvaldsdóttir$^2$ \\ \textbf{Mathias Müller}$^1$, \textbf{Sarah Ebling}$^1$ \\
  $^1$Department of Computational Linguistics, University of Zurich \\
  $^2$Communication Centre for the Deaf and Hard of Hearing in Iceland \\
  \texttt{finnuragust.ingimundarson@uzh.ch}\\
  \texttt{gudny.bjork.thorvaldsdottir@shh.is}\\
  \texttt{\{mmueller,ebling\}@cl.uzh.ch}}

\begin{document}

\maketitle

\begin{abstract}
We present the first experiments on isolated sign language recognition (ISLR) for Icelandic Sign Language (ÍTM). We use ÍTM SignWiki, a dataset derived from a bilingual Icelandic--ÍTM online dictionary. It is genuinely low-resource: 1,845 videos cover 849 classes, 86\% of which have only two examples, making the full task effectively one-shot recognition across signers. We compare two open-source ISLR frameworks, OpenHands and SPOTER, on three tasks of increasing vocabulary size (22, 117 and 849 classes), and evaluate three pose estimators and two forms of cross-lingual transfer. With ÍTM data alone, SPOTER outperforms OpenHands on all three tasks, and MediaPipe poses give better results than AlphaPose or SDPose. Cross-lingual transfer brings the largest gains: pretraining SPOTER on American Sign Language data before finetuning on ÍTM raises accuracy by 14--24 percentage points, to 72.7\%, 47.9\% and 22.6\% on the three tasks, and multilingual training with data from six other sign languages lifts OpenHands from 1.41\% to 28.86\% on the full task. Although far from practical use, the results suggest that transfer from better-resourced sign languages is promising for very low-resource ones. We release our adapted versions of both frameworks.
\end{abstract}

\section{Introduction}

Although not always recognized as such, sign languages are full-fledged natural languages that have their own grammar and vocabulary.\footnote{This paper is based on \citet{ingimundarson-islr-itm-real} the first author's master's thesis at the University of Zurich}
%and minimally extends it.}
Contrary to a widespread belief, there is no universal sign language and several hundred sign languages have been documented around the world. The number of deaf people who use them as a primary means of communication worldwide is estimated to be roughly 70 million, and more than half a million in Europe \citep{way_sl_mt_book}. 

The field of sign language processing (SLP) is more recent than that of natural language processing (NLP) and has lagged behind, constrained by a lack of technological or computational resources, software, and data \citep{yin-etal-2021-including,muller-etal-2022-findings}. This situation has gradually improved but SLP remains underrepresented and fundamental design decisions underexplored \citep{obrien_sant_pose_estimators}. In general, it can also be argued that all sign languages are low-resource languages when compared to spoken languages \citep{joshi-etal-2020-state}. Nevertheless, larger sign language communities, such as that of American Sign Language (ASL), are comparatively well represented and well resourced, whereas sign languages of smaller communities are low-resource, which restricts the development and deployment of SLP tools based on them. 

One such low-resource sign language is Icelandic Sign Language, that will hereafter be abbreviated as ÍTM (\textit{\textbf{í}slenskt \textbf{t}ákn\textbf{m}ál}), which is preferred by the Icelandic Deaf community and ÍTM researchers. ÍTM is now estimated to have about 300 native users, and in total approximately 2,000 users. The total number includes L2 users with late onset of hearing loss, children of deaf adults, foreign L1 signers (ÍTM L2 signers) and hearing signers with various levels of proficiency, the last group comprising somewhere between 1,000 and 1,500 people \citep{rannveig_koulidobrova_itm_bilingualism}. 

Automatic processing of ÍTM has not taken place so far and this paper represents one of the first steps in that direction. The task at hand, isolated sign language recognition (ISLR), is not the same as translation, and therefore less practical than a translation system.
%The continued narrow focus there has been on ISLR in recent years has been described as ``indicative of a lack of progress in the field and a lack of understanding of sign language and deaf needs'' \citep[p. 73]{FoxPractices}, as any application of it is of limited use to deaf people.
Nevertheless, for a low-resource sign language such as ÍTM, ISLR is a good starting point. With that in mind, this paper explores how well ISLR can perform in a real low-resource setting, on limited ÍTM data.

The contributions of this paper include:

\begin{enumerate}
    \item The first-ever ISLR experiments on ÍTM data
    \item Comparison of SPOTER and OpenHands, two open-source frameworks for ISLR
    \begin{enumerate}
        \item SPOTER yields significantly better baseline results than OpenHands on all three tasks, outperforming it by 36.36, 14.53 and 6.95 in test accuracy 
        \item Multilingual training in OpenHands improves the test performance by 27.8 and 27.45 percentage points
        \item Pretraining on ASL data and finetuning on ÍTM data improves SPOTER test performance by 22.73, 23.93 and 14.25 percentage points
        \item MediaPipe pose estimates in SPOTER performed better than AlphaPose and SDPose pose estimates  
    \end{enumerate}
\end{enumerate}

\section{Related Work}
\label{related_work}

Automatic sign language recognition (SLR) can be split into two tasks, \textit{continuous} or \textit{isolated} SLR \citep{koller2020quantitative}. In continuous SLR (CSLR), the recognition is performed either directly on the continuous stream or with an intermediate segmentation step, where sign boundaries are identified and the recognition is then performed on isolated signs based on the assumption that a single sign/gloss is contained within the segment boundaries. Isolated SLR (ISLR) is performed on single sign input.

The recognition is then treated as a classification task where information is extracted from signed video input that is processed into a representation suitable for a downstream task \citep{way_sl_mt_book}. The output labels can be, for instance, words or lemmas in the corresponding spoken language or sign language glosses.
%which could additionally be mapped to another sign language writing systems, such as HamNoSys or SignWriting.
%Examples of downstream tasks include searching for isolated signs in videos, lookup in sign language dictionaries, SLR as part of a translation pipeline or as part of sign language learning tools, where SLR can be applied to sign language assessment. 

\subsection{Video-based Recognition}
As observed by \citet{way_sl_mt_book}, in the wake of deep learning, the field of SLR has witnessed a surge of new techniques, models and model architectures. One of the first end-to-end approaches to SLR was the work of \citet{camgoz_2017_subunets}, who drew on speech recognition and the use of Connectionist Temporal Classification (CTC) algorithms. The novel architecture they proposed for sequence-to-sequence learning, called \textit{SubUNets}, consisted of three layers: a CNN layer to extract spatial features from image input, bidirectional LSTM layers to temporally model those spatial features, and a CTC loss layer on top.
%to handle differing video and label sequence lengths.
This was followed by the \textit{Sign Language Recognition Transformer} \citep{camgoz2020signlanguagetransformersjoint}, a unified model which was trained to jointly learn CSLR and translation.
%The encoder is trained to recognize glosses from continuous sign language videos, where spatial embeddings of video frames are obtained from a CNN, and the decoder is implemented as an auto-regressive translation model that uses the spatio-temporal representations learned by the encoder. This achieved state-of-the-art results for both recognition and translation, surpassing previous text-to-text translation results.

\subsection{Pose-based Recognition}
\label{subsec:pose-based-recognition}

Pose estimation provides a lower-dimensional alternative to raw video. Pose estimators extract human skeletal keypoints, creating signer-invariant representations that abstract over clothing, skin color, gender etc., focusing on the most relevant parts of the signal and ignoring irrelevant RGB features in the video. Despite these abstractions, the poses remain human-interpretable, but are not entirely anonymous \citep{battisti-etal-2024-person}.
Previous work has predominantly used two pose estimators: OpenPose \citep{cao_openpose} and MediaPipe \citep{lugaresi2019mediapipeframeworkbuildingperception}. Neither was developed specifically for SLP, but given their lightweight nature compared to raw videos, they have become widespread in SLP.

A recent and comprehensive comparison study of pose estimators is the work of \citet{obrien_sant_pose_estimators}. They compare eight different pose estimators and evaluate them on sign language translation (SLT). They show that four pose estimators achieve higher translation scores than MediaPipe, and suggest that alternatives to MediaPipe should be more strongly considered for pose-based SLT. Well-performing and interesting alternatives include Sapiens \citep{khirodkar_sapiens_pe} and SDPose \citep{liang2026sdposeexploitingdiffusionpriors} -- which are the best-performing estimators, but have significant compute requirements -- and AlphaPose \citep{alphapose}, which had the third-best translation performance and the fastest compute speed.

\subsection{Publicly Available ISLR Frameworks}
\label{subsec:publicly-available-islr-frameworks}

% SLR frameworks are seldom made publicly available  and OpenHands and SPOTER are among the few exceptions.

\paragraph{SPOTER} 

The \textit{Sign Pose-based Transformer} \citep{bohacek_spoter_transformer},\footnote{\color{blue}\href{https://github.com/maty-bohacek/spoter}{https://github.com/maty-bohacek/spoter}} proposed using pose estimates from Apple's Vision API to train a word-level SLR model. Along with a novel normalization scheme, they presented new data augmentation methods, namely four different spatial augmentations and adding Gaussian noise, which gave the best results. In addition, they demonstrated how well the SPOTER model performs on limited training data compared to an I3D model in the same setting. Their code and data availability has enabled follow-up works such as \citet{Azad_2026} who introduce several modifications to the original SPOTER, use MediaPipe poses instead of Vision API, adapt the implementation to Bengali Sign Language data, and substantially improve the previous results on an existing ISLR dataset. 

\citet{bohacek2022combiningefficientprecisesign} improves the original SPOTER approach by replacing the Vision API poses with MediaPipe Holistic.
%-- the subtitle of the paper is: \textit{Good pose estimation library is all you need}.
The poses were adapted to the landmarks used in the original SPOTER, in addition to performing hyperparameter search over augmentation parameters. By doing so, they greatly improved the performance on WLASL100, achieving state-of-the-art results for it at the time. This code is, however, not available.

SPOTER only implements one model architecture (Transformer) and neither pretraining strategies nor multilingual training setups are supported.

\paragraph{OpenHands}

Another important contribution in terms of open-sourced code is the OpenHands library by \citet{selvaraj-etal-2022-openhands},\footnote{\color{blue}\href{https://github.com/AI4Bharat/OpenHands}{https://github.com/AI4Bharat/OpenHands}} (see also \citet{nc2022addressing}) which aimed to exploit key low-resource language insights from traditional NLP for the benefit of sign languages. The open-source framework they released is no longer maintained, but includes pose estimates of existing datasets for five sign languages and more than 1,000 hours of Indian Sign Language data for self-supervised training.

OpenHands supports four model variants: two sequence-based models, RNN and Transformer, and two graph-based models, a spatio-temporal graph convolutional network (ST-GCN) and a sign language GCN (SL-GCN). Based on the results reported by \citet{selvaraj-etal-2022-openhands}, the graph-based models outperformed the sequence-based models on all of the datasets where accuracy was reported, with the SL-GCN performing best overall. 

%Additionally, they evaluate three different pre-training strategies, identifying Dense Predictive Coding as the most effective one, compared to masking and contrastive-learning based pre-training.
%
The framework also provides a multilingual training configuration that allows combining data from multiple sign languages. Two vocabulary strategies are available: a unified vocabulary, where the original classes from each dataset are ``normalized'' into English glosses, or a combined vocabulary of the original vocabulary of each dataset. 

In its default configuration, OpenHands supports 11 datasets for 7 sign languages (American, Chinese, Indian, Turkish, Greek, Argentinian and German).\footnote{See \href{https://openhands.ai4bharat.org/en/latest/instructions/datasets.html}{https://openhands.ai4bharat.org/en/latest/instructions/\\datasets.html} for an overview of the supported datasets.} However, three of these datasets are not publicly available. For the other datasets, the authors provide MediaPipe poses.

\section{Dataset}
\label{sec:dataset_methods}
We use the \textit{ÍTM SignWiki} dataset presented in \citet{ingimundarson-islr-itm-real}. It is based on the dictionary component of the Icelandic part of \textit{SignWiki},\footnote{\color{blue}\href{https://is.signwiki.org/index.php/Forsíða}{https://is.signwiki.org/}} a web and mobile platform for sign languages and deaf education. The dictionary is bilingual, where ÍTM sign language videos are mapped to Icelandic words or phrases, and currently contains roughly 13,000 signs and phrases. Although the dictionary is open for public input, the vast majority of the recordings originate from the Communication Centre for the Deaf and Hard of Hearing in Iceland (SHH), contributed by people who have been or are currently employed there, and the recordings date from the early 1990s and up until the present day.

\subsection{Dataset Profile}
\label{sec:corpus_profile}
An overview of the dataset is given in Tables \ref{tab:dataset_stats} and \ref{tab:samples_per_class}. As the numbers illustrate, the dataset is very limited in terms of size. Instead of having many examples on a limited set of classes, there are limited examples for many classes. This is in stark contrast to most ISLR datasets, such as the subsets of the MS-ASL dataset \citep{joze2019msasllargescaledataset}, that range from 100--1,000 classes and have, on average, 25.5--57.4 samples per class (189--222 signers).

\begin{table}
\centering
\small
\begin{tabular}{lr}
\toprule
%\textbf{} & \textbf{Count / Length} \\
%\midrule
Number of videos  & 1,845    \\
Total length (hours:minutes:seconds)  & 1:44:29 \\
Number of  classes & 849     \\
Number of signers & 33      \\
\bottomrule
\end{tabular}
\caption{Overall statistics of ÍTM SignWiki dataset}
\label{tab:dataset_stats}
\end{table}

\begin{table}
\small
\centering
\begin{tabular}{rrr}
\toprule
\textbf{\# Classes} & \textbf{\# Samples} & \textbf{Proportion} \\
\midrule
732 & 2 & 86.22\% \\
95 & 3  & 11.19\% \\
18 & 4  &  2.12\% \\
2 &  5  &  0.24\% \\
1 &  6  &  0.12\% \\
1 &  8  &  0.12\% \\
\bottomrule
\end{tabular}
\caption{Samples per class distribution in the ÍTM SignWiki dataset. Most classes represented by two samples}
\label{tab:samples_per_class}
\end{table}

\begin{figure*}[t]
     \centering
     \includegraphics[width=0.7\linewidth]{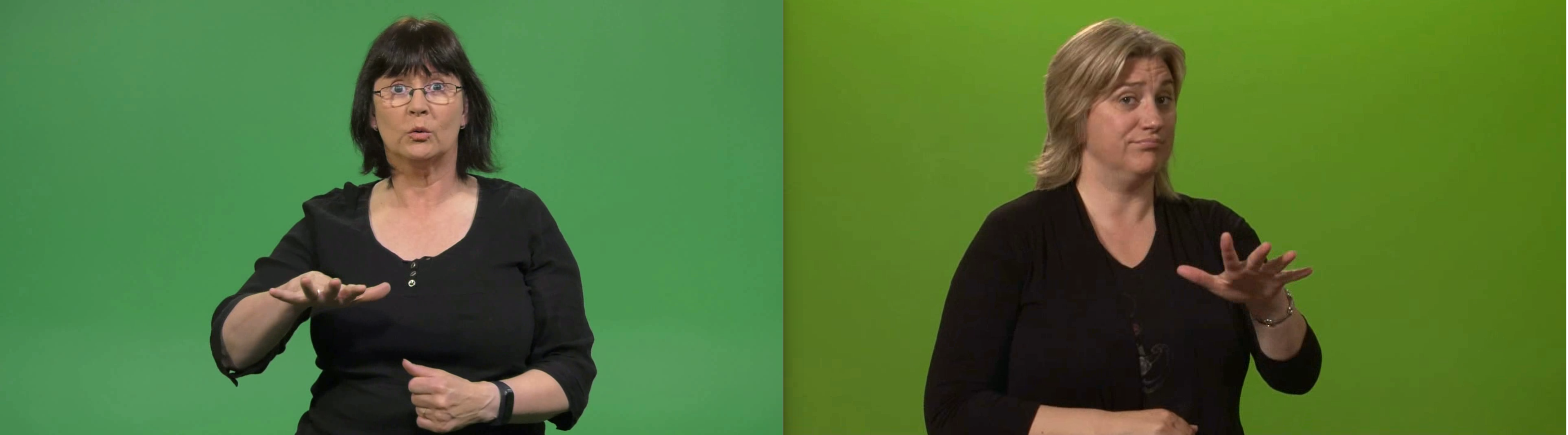}
     \caption{Example of left-/right-handed contrast in the ÍTM SignWiki dataset}
     \label{fig:dataset_example}
 \end{figure*} 

 As far as the signers in the data are concerned, 84.77\% are L1 deaf, 10.30\% are hearing and 4.93\% are L2 deaf. It is also worth noting that 89.7\% of the signers are female and 10.3\% are male, and all are white, which gives an indication of the overall composition of the data. As the data has been collected over many years, some of the more frequently featured signers have several different appearances, i.e.\ wearing different clothing, with different hairstyles etc.\ This would increase the diversity in a video-/RGB-based approach, but is less relevant for pose-based methods. 

A further characteristic of the dataset is the fact that the second-most frequently occurring signer in the dataset (Signer 1) is left-handed, the only one out of the 33 signers. The signer appears in 267 out of 1,845 samples. Therefore, many of the classes have a left-/right-hand contrast in addition to the signer difference between training and test samples. In total, 1,075 instances are labeled as right-handed, 202 left-handed, and 568 samples are two-handed symmetrical signs. An example of left-/right-handed contrast is shown in Figure \ref{fig:dataset_example}.

It is worth noting that a few instances of phrases, as opposed to isolated signs, are included in the data, e.g.\ \textit{hvað heitir þú} (`what is your name'). The dataset also contains multiple instances of other classes that consist of two or even three signs, such as the word \textit{sérkennari} `special needs teacher', composed of the signs SÉR (`special'), KENNA (`teach'), and PERSÓNA (`person'). Furthermore, seven of the 849 classes are fingerspelled. It must also be noted that the vocabulary of the dataset is unsystematic and contains, for example, signs for the words \textit{Algeria}, \textit{browned potatoes}, \textit{Lord} (in Christian sense), \textit{feather} and \textit{hippopotamus}. 

\subsection{Data splits and recognition tasks}
\label{material:data_splits}

\begin{table}
\small
\centering
\begin{tabular}{llrr}
\toprule
\textbf{Task} & \textbf{Split} & \textbf{\# Classes} & \textbf{\# Samples}\\
\midrule
\multirow{3}{*}{\textit{Minimal}}
& Train &    22   &   52  \\
 & Validation &   22    &   22  \\
 & Test &   22    &  22 \\
\midrule
\multirow{3}{*}{\textit{Trimmed}}
 & Train &   117    &  185  \\
 & Validation &   79    &  79  \\
& Test &   117   &  117  \\
 \midrule
 \multirow{3}{*}{\textit{Full}}
 & Train    & 849 & 917 \\
& Validation    & 79 & 79  \\
& Test    & 849 & 849 \\
\bottomrule
\end{tabular}
\caption{Recognition tasks based on the ÍTM SignWiki dataset, varying number of classes in the training, validation and test sets}
\label{tab:task_stats}
\end{table}

\paragraph{Data splits} We split the full dataset into a training, validation and test set with 917, 79 and 849 samples, respectively. The difference in size between the validation and test splits is due to dataset constraints. For 38 of the 117 classes that have three samples or more, two samples were retained for training instead of having a validation sample. In many cases, the validation sample is a different recording by the same signer as the training sample, whereas the test sample is drawn from a different signer not represented for that class. Therefore, the performance on the validation set might be misleading with regard to test set performance. This is preferable to validating on an unseen signer while testing on a seen signer for the same class.

\paragraph{Tasks} We define three different recognition tasks with increasing difficulty (varying random chance of success and class balance of training and validation data), see Table \ref{tab:task_stats}. The \textit{Minimal} task entails recognizing 22 classes (with $\geq$ 4 samples per class), the \textit{Trimmed} task has 117 classes (with $\geq$ 3 samples per class) and the \textit{Full} task has 849 classes (with $\geq$ 2 samples per class). Moving to the next task means adding more, and more challenging, classes, with the Full task effectively being one-shot recognition over 849 classes.

\section{Experiments}
We train a series of baselines with OpenHands and SPOTER using only the ÍTM SignWiki dataset (Section \ref{subsec:experiments-baselines}). Then we perform additional experiments on multilingual training (Section \ref{subsec:multilingual-training}) and a pretraining/finetuning scheme (Section \ref{subsec:pretraining-finetuning-scheme}) where the training data includes other languages as well.

\subsection{Baselines}
\label{subsec:experiments-baselines}

\paragraph{OpenHands}

We train systems for all four model variants that OpenHands supports: two sequence-based models, RNN and Transformer (BERT), and two graph-based models, a spatio-temporal graph convolutional network (ST-GCN) and a sign language GCN (SL-GCN) (see Section \ref{subsec:publicly-available-islr-frameworks}).

The framework provides precomputed poses for eight ISLR datasets (listed in Table \ref{tab:multilingual_original_results}), but also a pipeline to extract MediaPipe poses from video data. We used this pipeline on the ÍTM data. In the native OpenHands setup, the authors then define two different keypoint presets for the pose estimates: \textit{Minimal} (27 2D keypoints of the upper-body, hands and face) or \textit{Top body} (59 keypoints with better coverage of the upper body). Following \citet{selvaraj-etal-2022-openhands}'s example scripts we only used the minimal preset in all of our experiments.

In addition to trying different model architectures, we evaluate on all three recognition tasks (see Section \ref{material:data_splits}) and either disable or enable data augmentation. Taken together, we train $4\times3\times2 = 24$ baseline OpenHands models.

In the native OpenHands setup, the validation accuracy is monitored for early stopping, and we followed this example (max epochs=500/1000, patience=80 mode=max).
%In case of the full dataset, where the validation split has only 79 out of 849 classes, this is not ideal, but no efforts were made to monitor a different metric.
Further hyperparameters were the following: CosineAnnealingLR scheduler with Adam, lr = $1e-3$, and batch size 4/8/16 for \textit{Minimal}/\textit{Trimmed}/\textit{Full}.
Models were trained on an A100 or H100 GPU (this is true for all models in our experiments).

\paragraph{SPOTER}
The best results with SPOTER were achieved with MediaPipe poses as the input representation (see Section \ref{subsec:publicly-available-islr-frameworks}). For this paper, we adapted the SPOTER code to the binary pose format developed by \citet{moryossef2021pose-format}. 
Inspired by \citet{obrien_sant_pose_estimators} we compare three different pose estimators: AlphaPose \citep{alphapose}, MediaPipe \citep{lugaresi2019mediapipeframeworkbuildingperception} and SDPose \citep{liang2026sdposeexploitingdiffusionpriors}, instead of assuming MediaPipe as a fixture of the experiment. 

As described in Section \ref{related_work}, MediaPipe and OpenPose have been the two predominant pose estimators used in SLP in recent years. AlphaPose and SDPose are lesser known in an SLP context. In \citet{obrien_sant_pose_estimators}'s comparison of pose estimators, evaluated on SLT, AlphaPose ranked third, and was found to be more computationally efficient than the better-performing estimators. It is primarily the low-latency benefit of AlphaPose that makes it appealing for the purposes of this project. While working with a low-resource language does not necessarily entail limited computational resources, a lightweight solution with competitive performance would be advantageous.

% See Section \ref{subsec:pose-based-recognition} for an extended rationale for why trying alternative pose estimators is promising.

% We chose SDPose because it achieved the highest BLEU score in translation experiments \citep{obrien_sant_pose_estimators} -- tied with Sapiens -- but it was also among the most resource-intensive and slowest estimators in the evaluation. AlphaPose was a viable alternative. 

To extract the poses, we used the \textit{video-to-pose} repository by \citet{obrien-et-al-2026video-to-pose},\footnote{\color{blue}\href{https://github.com/ZurichNLP/video-to-pose}{https://github.com/ZurichNLP/video-to-pose}} which at the time of writing supports eight pose estimators in total.
In the SPOTER implementation, a total of 54 body landmarks are extracted, including five head landmarks (eyes, ears, and nose) and 21 body landmarks that represent body joints. This results in 54 2D points and an 108 dimensional feature vector for each frame.
Following \citet{bohacek_spoter_transformer} we train for 350 epochs and add Gaussian noise to the training set (along with other data augmentation techniques the framework provides).

To summarize, we trained SPOTER baselines for all three recognition tasks (see Section \ref{material:data_splits}) and for three different pose estimators ($3\times3 = 9$ baseline SPOTER models).

\subsection{Multilingual training}
\label{subsec:multilingual-training}

We train additional OpenHands models on more datasets in other sign languages (see Table \ref{tab:multilingual_original_results} in Appendix \ref{sec:appendix} for the full list of datasets).

\paragraph{Vocabulary strategies} We experiment with either keeping the vocabularies of all datasets separate (\textit{original}) or unifying them into a single, normalized vocabulary (\textit{unified}). In the first setting, the language code (ISO) of each sign language included is prepended to the class label as a one-hot encoded vector, i.e.\ \textit{ice}\_\_ for ÍTM.

In the second setting, the vocabulary of each non-English/non-ASL dataset is normalized to English glosses. Here, the ÍTM subset has only 817 classes as the normalization allows to combine different sign variants of the same sign, of which there are several examples in the dataset, with the same normalized gloss. See Appendix \ref{sec:appendix} for a more detailed explanation.

This results in a single normalized vocabulary and predictions are therefore made with normalized glosses, which can then be mapped back to the original vocabulary, but not to separate variants. As the unified vocabulary experiment effectively is a different version of the \textit{Full} task, we keep those results separate and present them in Appendix \ref{sec:appendix}.

These experiments use the multilingual training feature of OpenHands, described in \citet{nc2022addressing}, but only the unified vocabulary, and the experiments were inspired by example configs for multilingual training provided with the framework.\footnote{See \color{blue}\href{https://github.com/AI4Bharat/OpenHands/tree/main/examples/configs/multilingual}{https://github.com/AI4Bharat/OpenHands/tree/main/-examples/configs/multilingual}} We used the following hyperparameters: CosineAnnealingLR scheduler with Adam, learning rate $1e-3$, batch size 64, max epochs 200, with early stopping on validation accuracy (max, patience 30).

Due to time and resource constraints, we evaluate these multilingual models only on the \textit{Full} recognition task.

\subsection{Pretraining/finetuning scheme}
\label{subsec:pretraining-finetuning-scheme}

SPOTER does not support pretraining out-of-the-box. For this paper, we adapted the framework to support pretraining (either with the encoder frozen for a specified number of epochs or unfrozen from the start) and finetuning.

For this experiment, a SPOTER model was pre-trained on the ASL Citizen dataset \citep{asl_citizen}. It is a community-sourced dataset for ASL that has nearly 84,000 videos filmed by 52 signers and 2,731 classes. One motivation for using it was that MediaPipe poses in pose format for the dataset were already available, though any comparable ISLR dataset for a higher-resource sign language would have been a viable alternative. As MediaPipe outperformed the other two pose estimators in the baseline experiments (see Section \ref{subsec:baselines}), this setup was exclusively tested with MediaPipe poses. As the aim here was simply for the encoder to learn general representations of ASL and not to achieve the best results on the dataset, the training was limited to 30 epochs (test accuracy 40.28).

For the finetuning we tested two approaches: one where the encoder was frozen for the first 30 epochs and only the decoder and head trained, and another where the full model was finetuned from the beginning.

Thus the model is pre-trained on ASL data and finetuned on ÍTM, with the encoder either frozen from the start or unfrozen; exclusively with MediaPipe poses ($3\times2$ experiments).

\subsection{Evaluation Metrics}

We report test accuracy and validation accuracy, given that the validation set is somewhat particular and limited in two ways (see Section \ref{material:data_splits}): on the one hand it covers only 79 classes (compared to the total of 849) and on the other hand, the validation samples are mostly with the same signer as (one of) the training sample(s), whereas the signer in the test set is always unseen for that particular sign.
%Therefore, the difference between the validation and test set accuracy can, at least in part if not entirely, be explained by these facts.

\section{Results}
\label{sec:results}

This section presents the initial ISLR experiments we performed with the two frameworks. An overview of the best results is shown in Table \ref{tab:results_overview} and an additional table for the multilingual results is included in Appendix \ref{sec:appendix}.

\subsection{Baselines (ÍTM Data Only)}
\label{subsec:baselines}

\paragraph{OpenHands}

\begin{table}
\scriptsize
\centering
\begin{tabular}{llcrr}
\toprule
\textbf{Task} & \textbf{Model} & \textbf{Augmentation} & \textbf{Validation} & \textbf{Test} \\
\midrule
\multirow{8}{*}{\textit{Minimal}}
& \multirow{2}{*}{LSTM}        & \checkmark & 18.20 & 9.09 \\
&                              & -          & 9.1   & 0.00 \\
\cmidrule(l){2-5}
& \multirow{2}{*}{Transformer} & \checkmark & 22.70 & 9.09 \\
&                              & -          & 31.8  & 9.09 \\
\cmidrule(l){2-5}
& \multirow{2}{*}{ST-GCN}      & \checkmark & 13.60 & 9.09 \\
&                              & -          & 40.9  & \textbf{13.64} \\
\cmidrule(l){2-5}
& \multirow{2}{*}{SL-GCN}      & \checkmark & 9.10  & 4.54 \\
&                              & -          & 54.5  & 9.09 \\
\midrule
\multirow{8}{*}{\textit{Trimmed}}
& \multirow{2}{*}{LSTM}        & \checkmark & 3.80  & 0.00 \\
&                              & -          & 2.5   & 0.85 \\
\cmidrule(l){2-5}
& \multirow{2}{*}{Transformer} & \checkmark & 3.80  & 1.70 \\
&                              & -          & 7.6   & 2.56 \\
\cmidrule(l){2-5}
& \multirow{2}{*}{ST-GCN}      & \checkmark & 21.5 & 2.56 \\
&                              & -          & 51.9  & 3.42 \\
\cmidrule(l){2-5}
& \multirow{2}{*}{SL-GCN}      & \checkmark & 10.10 & 4.27 \\
&                              & -          & 59.5  & \textbf{9.40} \\
\midrule
\multirow{8}{*}{\textit{Full}}
& \multirow{2}{*}{LSTM}        & \checkmark & 7.60  & 0.00 \\
&                              & -          & 10.1  & 0.23 \\
\cmidrule(l){2-5}
& \multirow{2}{*}{Transformer} & \checkmark & 2.50  & 0.00 \\
&                              & -          & 3.8   & 0.11 \\
\cmidrule(l){2-5}
& \multirow{2}{*}{ST-GCN}      & \checkmark & 22.8 & 0.71 \\
&                              & -          & 26.6  & 1.06 \\
\cmidrule(l){2-5}
& \multirow{2}{*}{SL-GCN}      & \checkmark & 17.70 & \textbf{1.41} \\
&                              & -          & 30.4  & 0.82 \\
\bottomrule
\end{tabular}
\caption[OpenHands: ÍTM Data Only Results]{Accuracy of OpenHands baseline models trained on ÍTM data only (\checkmark=with augmentation, -=only normalization)}
\label{tab:openhands-baseline}
\end{table}

Table \ref{tab:openhands-baseline} shows the performance of all OpenHands baselines. The graph-based models generally outperform the LSTM and Transformer models, with a GCN variant achieving the best test accuracy of 13.64\% / 9.40\% / 1.41\% on the \textit{Minimal} / \textit{Trimmed} / \textit{Full} tasks respectively. Only one configuration yielded more than 10\% accuracy on the \textit{Minimal} task; an ST-GCN model without any data augmentation. For reference, the best OpenHands baseline model is repeated in Table \ref{tab:results_overview}.

Across the baselines, disabling augmentation yields slightly better test accuracy in 8 of 12 pairs and in general higher validation accuracy. However, since the test sets are rather small (22, 117 and 849 examples) these are in fact minor differences.

\begin{table}
\small
\centering
\begin{tabular}{llrr}
\toprule
\textbf{Task} & \textbf{Poses} & \textbf{Validation} & \textbf{Test}\\
\midrule
\multirow{3}{*}{\textit{Minimal}}
& MediaPipe &    72.72   &   \textbf{50.00}  \\
 & AlphaPose &   63.63    &   45.45  \\
 & SDPose &   63.63    &  31.82 \\
\midrule
\multirow{3}{*}{\textit{Trimmed}}
 & MediaPipe &   72.15    &  \textbf{23.93}  \\
 & AlphaPose &   59.49    &  18.80  \\
& SDPose &   59.49   &  17.09  \\
 \midrule
\multirow{3}{*}{\textit{Full}}
 & MediaPipe    & 67.09 & \textbf{8.36} \\
& AlphaPose    & 53.16 & 3.06  \\
& SDPose    & 55.70 & 4.00 \\
\bottomrule
\end{tabular}%
\caption{Accuracy of SPOTER baseline models trained on ÍTM data only, varying the pose estimation system}
\label{tab:spoter_itm}
\end{table}

\paragraph{SPOTER}
Table \ref{tab:spoter_itm} shows the performance of all SPOTER baselines. MediaPipe poses consistently outperforms other estimators, for example the test accuracy of MediaPipe is roughly 5 percentage points higher than AlphaPose on all three tasks. In general, the SPOTER baselines show considerably higher accuracy on the test set than comparable OpenHands baselines (see above).
%This discrepancy will be discussed in Section \ref{sec:discussion}.
For reference, the best SPOTER baseline model (only MediaPipe results) is repeated in Table \ref{tab:results_overview}.

\subsection{Multilingual Training}
\label{sec:openhands_multilingual}

Multilingual training results with the original vocabulary are shown in Table \ref{tab:results_overview}, for a direct comparison with baseline scores. Individual per-dataset scores are reported in Table \ref{tab:multilingual_original_results} in Appendix \ref{sec:appendix}. Multilingual training results only concern the \textit{Full} task. Adding multilingual training outperforms the best OpenHands and SPOTER baselines. For example, the test accuracy of the best SPOTER baseline is 8.36, while the test accuracy of the best multilingual model is 28.86. Unifying the multilingual vocabulary (as opposed to keeping separate vocabularies) yields higher accuracy, 29.21, but on a slightly smaller vocabulary (see Section \ref{subsec:appendix_multilingual}).

\subsection{Pretraining / finetuning scheme}

\begin{table}
\small
\centering
\begin{tabular}{llrrr}
\toprule
\textbf{Task} & \textbf{Setting} & \textbf{Val}& \textbf{Test}& \textbf{$\Delta$Baseline} \\
\midrule
\multirow{2}{*}{\textit{Minimal}}
& Frozen & 95.45 & \textbf{72.73} & +22.73\\
& Unfrozen& 95.45 & 59.09 & +9.09 \\
 \midrule
 \multirow{2}{*}{\textit{Trimmed}}
  & Frozen & 82.27 & \textbf{47.86} & +23.93 \\
 & Unfrozen & 67.09 & 26.50 & +2.57 \\
\midrule
\multirow{2}{*}{\textit{Full}}
 & Frozen    & 78.48 & \textbf{22.61} & +14.25 \\
 & Unfrozen  & 68.35 & 13.43 & +5.07 \\
 \bottomrule
\end{tabular}%
\caption{Test accuracy of a SPOTER model pre-trained on ASL and finetuned on ÍTM compared to the baseline in Table \ref{tab:spoter_itm}  (Frozen=Pre-trained encoder frozen for first 30 epochs, Unfrozen=Full model finetuned from start)}
\label{tab:ASL_itm_finetune}
\end{table}

\begin{table*}
\small
\centering
\begin{tabular}{lrlllrrr}
\toprule
\textbf{Task} & \textbf{Random} & \textbf{Framework} & \textbf{Model} & \textbf{Setting} &  \textbf{Val}&\textbf{Test} &\textbf{Time}\\
\midrule
\multirow{4}{*}{\textit{Minimal}} & \multirow{4}{*}{4.54} & OpenHands & ST-GCN & Baseline (No Augmentation)&  40.9&13.64 & 00:01:54\\
\cmidrule{3-8}
& & SPOTER & Transformer & Baseline (MP)&  72.72&50.00  & 00:02:18\\
& & SPOTER & Transformer & ASL-Finetuned (MP)-Frozen & 95.45&\textbf{72.73} &*05:22:24\\
 
\specialrule{1pt}{2pt}{2pt}
 \multirow{4}{*}{\textit{Trimmed}} & \multirow{4}{*}{0.85} & OpenHands & SL-GCN& Baseline (No Augmentation)&  59.5 & 9.40 & 00:07:39\\
\cmidrule{3-8}
& & SPOTER & Transformer & Baseline (MP)&  72.15&23.93  &00:18:06\\
& & SPOTER & Transformer & ASL-Finetuned (MP)-Frozen   &  82.27&\textbf{47.86}  &*05:25:13\\
\specialrule{1pt}{2pt}{2pt}

\multirow{5}{*}{\textit{Full}} & \multirow{5}{*}{0.12} & OpenHands & SL-GCN & Baseline (Augmentation)&  17.70&1.41 & 01:14:32\\
& & OpenHands & SL-GCN & Multilingual Original  &  68.2&\textbf{28.86} & 08:15:16 \\
\cmidrule{3-8}
& & SPOTER & Transformer & Baseline (MP)&  67.09&8.36  &01:31:25\\
& & SPOTER & Transformer & ASL-Finetuned (MP)-Frozen   &  78.48&22.61  &*06:07:20\\
\bottomrule
\end{tabular}
\caption[Results Overview]{Validation and test accuracy on the three ÍTM recognition tasks for OpenHands and SPOTER. (Baseline=best baseline score, Random=Random classification accuracy, MP=MediaPipe Holistic poses, Time=Training time as HH:MM:SS, *=Combined training time (ASL Citizen pretraining took 05:21:11 hours)}
\label{tab:results_overview}
\end{table*}

% SPOTER

Pretraining / finetuning results are shown in full in Table \ref{tab:ASL_itm_finetune}. Pretraining on ASL data also outperforms the best baselines by at least 10 percentage points in test accuracy. For instance, on the \textit{Full} task, the best SPOTER baseline achieves 8.36 test accuracy, while the best finetuned model achieves 22.61 accuracy. Furthermore the results demonstrate that freezing the pre-trained encoder at the beginning of finetuning increases the test accuracy by at least 10 percentage points.

\section{Discussion}
\label{sec:discussion}

\paragraph{Baselines: SPOTER vs.\ OpenHands}

When only using ÍTM data, SPOTER clearly outperforms Openhands on all three tasks: 50.00 vs.\ 13.64, 23.93 vs.\ 9.40 and
8.36 vs.\ 1.41 test accuracy (see Section \ref{subsec:baselines}). These margins should be read with the size of the test sets in mind. On the \textit{Minimal} task each prediction is worth $1/22=4.54$ percentage points, and the gap amounts to 11 vs.\ 3 correct predictions; on \textit{Trimmed} it is 28 vs.\ 11 out of 117. The \textit{Full} task, with 71 vs.\ 12 correct predictions out of 849, therefore carries most of the evidence, and the smaller tasks should not be over-interpreted.

We emphasize that this is a comparison of two frameworks as they are distributed, not of two architectures. Input representation, normalization, augmentation, optimization and checkpoint selection all differ at the same time, and our experiments do not isolate these factors. We therefore offer hypotheses rather than explanations.

For example, OpenHands and SPOTER reduce the full set of MediaPipe keypoints in different ways. OpenHands offers 27-point and 59-point presets, but our OpenHands models use only the 27 point preset, following \citet{selvaraj-etal-2022-openhands}. SPOTER, on the other hand, uses 54 2D keypoints. This difference in keypoint resolution may in part explain the difference in performance between the frameworks: a coarser hand representation could plausibly mean a disadvantage for OpenHands. Re-running OpenHands with the 59-point preset would test this directly.

Second, the frameworks normalize differently. OpenHands applies a single shoulder-referenced centering and scaling to the whole skeleton, so hand landmarks occupy a small region of the normalized space, whereas SPOTER normalizes body and hands separately, distorting hand keypoints to a lesser degree. Normalization is intimately tied to generalization; because the test signer is always unseen for a given class (Section \ref{material:data_splits}), our test set specifically measures signer-invariant generalization. Consistent with this, SPOTER retains a much larger share of its validation accuracy on the test set (69\%/33\%/12\% across the three tasks) than the best OpenHands models do (33\%/16\%/8\%). Both frameworks overfit to the seen signer; OpenHands does so considerably more. The importance of preprocessing choices of this kind for pose-based SLP has been noted before \citep{decoster2023extractionrobustsignembeddings,obrien_sant_pose_estimators}.

Third, the training and model selection protocols differ. SPOTER trains for a fixed number of epochs, saves the two checkpoints with highest training and validation accuracy every ten epochs, and evaluates over all of them. Our OpenHands runs use early stopping and checkpoint selection on validation accuracy, which on the \textit{Full} task is computed over 79 of 849 classes, and a single checkpoint was then chosen manually from $k$ saved checkpoints. Model selection is thus both noisier and more weakly related to the target task for OpenHands, and we did not evaluate all saved checkpoints for comparison.

Fourth, and in our view most informative, the graph-based models appear to be data-starved rather than unsuited to the task. The same SL-GCN that reaches 1.41 on the \textit{Full} task with ÍTM data alone reaches 28.86 once other sign languages are added (Section \ref{sec:openhands_multilingual}). \citet{selvaraj-etal-2022-openhands} report their graph models performing best on datasets with tens of samples per class; ÍTM SignWiki offers one or two. In this respect our results are in line with \citet{bohacek_spoter_transformer}, who show SPOTER learning effectively from small training sets, although their comparison was against a video-based I3D model that must first learn general properties of human motion, whereas both frameworks compared here operate on poses. Our results are therefore consistent with their claim but do not test it under the same conditions.

Finally, although both frameworks use MediaPipe, the poses were extracted with different pipelines: the extraction script shipped with OpenHands in one case and \textit{video-to-pose} \citep{obrien-et-al-2026video-to-pose} in the other. Differences in version or configuration cannot be ruled out as a contributing factor.

\paragraph{SPOTER: Choice of pose estimator}

MediaPipe clearly outperformed AlphaPose and SDPose. It is worth reiterating that this was not a comparison of the full pose estimates but of the customized (reduced) SPOTER format. However, the format was the same for all three estimators and the training schemes identical. As mentioned in Section \ref{related_work}, a comparison study of pose estimators performed by \citet{obrien_sant_pose_estimators} revealed that three other pose estimators performed better than MediaPipe on SLT. In our experiments, the downstream task is different, recognition instead of translation, and the set of keypoints is reduced. Nevertheless, MediaPipe appears better suited to ISLR. This is in line with findings of the comparative studies of \citet{decoster2023extractionrobustsignembeddings} and \citet{moryossef2021evaluatingimmediateapplicabilitypose} of pose estimators for SLR, where MediaPipe outperformed OpenPose and MMPose. Our results are, to the best of our knowledge, the first comparison of MediaPipe, AlphaPose and SDPose for SLR.

\paragraph{Cross-lingual transfer effects}

Both multilingual training and fine-tuning a pretrained model improved the performance, meaning that both constitute genuine cross-lingual transfer effects from higher-resourced sign languages to a low-resource one.
Multilingual training raises the best OpenHands result on the \textit{Full} task from 1.41 to 28.86, a twentyfold increase, while ASL pretraining raises the best SPOTER result from 8.36 to 22.61, and by 14--24 percentage points across the three tasks. On the \textit{Full} task the multilingual model is ahead (245 vs.\ 192 correct predictions of 849).

These two numbers should not be read as a ranking of the two strategies. They come from different frameworks, whose baselines already differ by a factor of six; from different source data, eight datasets in six sign languages in one case and a single ASL dataset in the other; and from different label spaces at inference. The multilingual model predicts over all 5,732 classes of the concatenated dataset, and 137 of its 849 test predictions fall on classes belonging to other sign languages (see Appendix \ref{appendix:multilingual_predictions}); these are incorrect by construction, so its ÍTM accuracy is measured under a handicap that the pretrained model does not face. A controlled comparison would require training SPOTER jointly on ASL Citizen and ÍTM and evaluating both schemes within one framework, and the multilingual configuration on the two smaller tasks, neither of which we were able to do.

\section{Conclusion}

In this paper we explore the performance of two publicly available SLR frameworks in a low-resource setting, using the first ISLR dataset for ÍTM. The dataset is very limited with regard to the intended use in a machine learning task; it has many classes (849) and few samples per class, only two samples for the majority of classes (732). The experiments therefore explored how well SLR can perform in a real low-resource setting.

We demonstrate that multilingual learning and cross-lingual transfer can benefit lower-resource sign languages. In the OpenHands experiments, the multilingual training scheme achieved the best results on the full dataset. In the case of SPOTER, we show that SPOTER learns effectively from very small training sets, corroborating the findings of \citet{bohacek_spoter_transformer}. Additionally, finetuning on ÍTM after pretraining on ASL Citizen data yielded the best overall results on the original ÍTM vocabulary on two out of three tasks and demonstrates the effectiveness of a strong pre-trained checkpoint that can be finetuned to different tasks. The multilingual model with original vocabulary in OpenHands yields the best performance on the full (most challenging) task.

The repositories of the adapted versions of the two frameworks have been made publicly accessible on GitHub, in the hope that they may prove useful to other researchers,\footnote{\href{https://github.com/ZurichNLP/openhands-itm}{https://github.com/ZurichNLP/openhands-itm}}$^,$\footnote{\href{https://github.com/ZurichNLP/spoter-pose}{https://github.com/ZurichNLP/spoter-pose}} and the dataset will soon be published as well. 

\section{Limitations}
\label{limitations}

\paragraph{Framework choice} Both of these frameworks are from 2022 and comparison with at least one newer ISLR method would have been preferable, but no more recent, publicly available code could be found, except extensions of SPOTER.

\paragraph{Error analysis} Beyond general accuracy measures, we present no error analysis. Given the composition of the dataset, it could for instance be insightful to analyze errors based on sign types, e.g. one-handed vs. two-handed, fingerspelled signs and multi-sign compounds (see Section \ref{sec:corpus_profile}), or per-signer performance. 

\paragraph{Baseline tuning} The discrepancy in performance in the OpenHands baselines with regard to data augmentation, where the performance without augmentation was in general better, was unexpected and would need further inspection. The augmentation techniques used followed the examples of the OpenHands authors, but could perhaps be adjusted better to the ÍTM dataset.
%, given its composition compared to the considerably bigger ISLR datasets experimented with by the OpenHands authors.

\paragraph{Variability} We only present one single run for each training configuration. Training several models with different random seeds would make our results and conclusions drawn from them more robust.

\paragraph{Applicability} Although the experiments yielded meaningful results it should nonetheless be stressed that the results do not yet constitute practically applicable performance. ISLR systems can be applied to tasks such as looking for signs in videos or dictionaries, but none of the models we trained could be immediately deployed in such a setting. This would require substantially greater amounts of training data as well as a larger vocabulary.

\section{Acknowledgements}
We would like to thank three anonymous reviewers for useful comments. MM received funding from the SIGMA project (grant no. G-95017-01-07), supported by the Digital Society Initiative (DSI) at the University of Zurich.

The language of the abstract and discussion sections was refined with a Claude agent.


\bibliography{custom}

\newpage

\appendix
\section{Multilingual Results}
\label{sec:appendix}

\begin{table*}
\small
\centering
\begin{tabular}{lrrrrrr}
\toprule
& \multicolumn{2}{c}{\textbf{Original}} & \multicolumn{2}{c}{\textbf{Unified}} & \multicolumn{2}{c}{\textbf{OpenHands}} \\
\cmidrule(lr){2-3} \cmidrule(lr){4-5} \cmidrule(lr){6-7} 
\textbf{Dataset} & \textbf{ST-GCN} & \textbf{SL-GCN} & \textbf{ST-GCN} & \textbf{SL-GCN} & \citet{selvaraj-etal-2022-openhands}  &\citet{nc2022addressing}\\
\midrule
ASLLVD          & 50.92 &  \textbf{55.64} &  49.49 &55.38 &- &50.1\\
AUTSL           & 90.89 &  91.42 &  90.65 &91.47 & 91.9 &\textbf{92.1}\\
GSL             & 94.94 &  \textbf{96.08} &  94.40 &95.62 &95.4 &93.6\\
INCLUDE         & 96.07 &  96.57 &  94.85 &\textbf{97.55} &93.5 &96.3\\
LSA64           & 96.56 &  \textbf{98.75} &  96.56 &96.87 &97.8 &97.5\\
MSASL           & 62.20 &  65.89 &  61.48 & 66.44 &- &\textbf{67.4}\\
RWTH            & 0.41 &   1.03 &  0.20 &0.82 &- &\textbf{48.5}\\
WLASL           & 44.27 &  \textbf{47.43} &  44.92 &47.01 &30.6 &46.6\\
\midrule
\textbf{ÍTM (\textit{Full)}}\tnote{$\dagger$}      & 20.84 &  28.86 &  22.12 &\textbf{29.21} &- &-\\
\bottomrule
\end{tabular}
\caption{Per-dataset test accuracy for two multilingual models, 1) original vocabulary (5,732 classes) and 2) unified vocabulary (4,261 classes), along with the best standalone results (all SL-GCN) reported by \citet{selvaraj-etal-2022-openhands}}
\label{tab:multilingual_original_results}
\end{table*}

\subsection{Multilingual Training}
\label{subsec:appendix_multilingual}
In Table \ref{tab:multilingual_original_results}, we present the results of the test set evaluation for both vocabulary approaches, unified and original, of the multilingual training, with two different architectures, along with a comparison of the results reported by \citet{selvaraj-etal-2022-openhands} and \citet{nc2022addressing}. 

The multilingual training outperforms the best accuracy reported by \citet{selvaraj-etal-2022-openhands} and \citet{nc2022addressing} on five datasets, ASLLVD, GSL, INCLUDE, LSA64 and WLASL, and in four cases out of five with the original vocabulary and an SL-GCN. The discrepancy in performance on the RWTH-PHOENIX dataset is notable. This data differs from the other datasets, as the poses have been extracted from low-resolution image files instead of videos that are furthermore from continuous data. 

Our setup is not completely comparable to the work of \citet{nc2022addressing}, where the unified vocabulary approach is used. We include the ÍTM dataset, whereas they include -- in addition to the datasets we use -- one Turkish Sign Language and two Chinese Sign Language datasets, that were not available to us. They explore different pretraining and finetuning options: one where the unified vocabulary (referred to as \textit{MultiSign-ISLR dataset} is pretrained on Indian Sign Language data and finetuned individually, another model pretrained on \textit{SignCorpus}, a large pretraining dataset of 4.6K hours of data from 10 different sign languages, and finetuned individually, and lastly pretrained on \textit{SignCorpus} and finetuned jointly on all of the sign languages.

\paragraph{Original vs. unified vocabulary}

As an alternative to a simple concatenation of all sign classes from all languages, we consider a \textbf{unified} vocabulary. As an example from our dataset, two different classes for sign variants of \textit{allt fínt} `all good' are mapped to the same normalized gloss, ALL\_GOOD. Including ÍTM in this experiment is admittedly only synthetic, as the original data is not glossed nor are there English glosses available. To synthesize this, the classes (words) in the ÍTM data were machine-translated to English, manually reviewed, and compared to normalized glosses for the other datasets -- with 40.8\% overlap between the translated ÍTM glosses and the other normalized glosses (see \citet{nc2022addressing} for their approach to the normalization).

Our results demonstrate how the unified vocabulary outperforms the original vocabulary and demonstrate the effectiveness of crosslingual transfer. It must, however, be stressed that the unified vocabulary has fewer labels than the original vocabulary since several classes that have variants in the dataset are collapsed into a single class in the unified vocabulary. The evaluation is therefore done on a test set with 817 classes instead of the full 849, as in the original vocabulary. And consequently, the results of the two vocabulary approaches are not fully comparable and the difficulty of the tasks is a confounding variable. 

In a similar vein, regarding the unified vocabulary approach, we emphasize that these results are based on synthetic ÍTM glosses that have not been verified by an ÍTM expert. Therefore, the results are only an indication of the benefits of this approach, whereas the results with the original vocabulary provide more direct evidence of the benefits of multilingual training. 

\section{Multilingual Prediction Analysis}
\label{appendix:multilingual_predictions}

If the original vocabulary is preserved in the multilingual OpenHands model, then the recognition is performed over a large multilingual label space, and predictions in other languages are possible.
On the full test set there are 137 cases of another sign language being predicted and the overall distribution is shown in Table \ref{tab:multilingual_sl_distribution}. This distribution is in line with the number of classes in the included datasets, and ASL has the largest part of the vocabulary in the concatenated dataset. It is worth considering whether these foreign predictions are in fact correct for the other sign languages. As an example of this, one might for instance look at the prediction for the ÍTM class \textit{frumskógur} `jungle', which is the plural form of tree in ASL (ase\_\_tree\_pl), or the ASL prediction \textit{boy} for the ÍTM class `man'. The similarity at the level of the word label or gloss alone suggests that the ASL predictions may be correct. However, this might also simply be a coincidence and there are multiple examples of foreign predictions with no label similarity to the ÍTM class, although the underlying signs may nonetheless be phonologically similar. This would require a closer inspection and comparison of the videos.

\begin{table}[h]
\centering
\small
\begin{tabular}{llr}
\toprule
\textbf{ISO} & \textbf{SL} & \textbf{Count}  \\
\midrule
ase & American & 88 \\
tsm & Turkish & 16 \\
gsg & German & 1 \\
ins & Indian & 20 \\
gss & Greek & 9 \\
aed & Argentinian & 3 \\
icl & Icelandic & 712 \\
\bottomrule\end{tabular}\caption[Multilingual SL predictions on ÍTM]{Multilingual test prediction distribution with original vocabulary (ST-GCN)}
\label{tab:multilingual_sl_distribution}
\end{table}

\end{document}